\pdfoutput=1

\documentclass[11pt]{article}

\usepackage[preprint]{acl}

\usepackage{times}
\usepackage{latexsym}

\usepackage[T1]{fontenc}

\usepackage[utf8]{inputenc}

\usepackage{microtype}

\usepackage{inconsolata}

\usepackage{graphicx}

\usepackage{booktabs}
\usepackage{siunitx}
\usepackage{makecell}

\title{Detecting Spelling and Grammatical Anomalies in Russian Poetry Texts}

\author{Ilya Koziev \\
  SalutDevices \\
  \small{
    \textbf{Correspondence:} \href{inkoziev@gmail.com}{inkoziev@gmail.com}
  }
}

\begin{document}
\maketitle
\begin{abstract}

The quality of natural language texts in fine-tuning datasets plays a critical role in the performance of generative models, particularly in computational creativity tasks such as poem or song lyric generation. Fluency defects in generated poems significantly reduce their value. However, training texts are often sourced from internet-based platforms without stringent quality control, posing a challenge for data engineers to manage defect levels effectively.

To address this issue, we propose the use of automated linguistic anomaly detection to identify and filter out low-quality texts from training datasets for creative models. In this paper, we present a comprehensive comparison of unsupervised and supervised text anomaly detection approaches, utilizing both synthetic and human-labeled datasets. We also introduce the \texttt{RUPOR} dataset, a collection of Russian-language human-labeled poems designed for cross-sentence grammatical error detection, and provide the full evaluation code. Our work aims to empower the community with tools and insights to improve the quality of training datasets for generative models in creative domains.
\end{abstract}

\section{Introduction}

The development of systems for automatic poem and song lyric generation involves addressing two critical challenges.

First, training transformer-based language models, which are widely used in computational creativity nowadays, requires large-scale datasets. The linguistic quality of these datasets significantly impacts the performance of downstream creative tasks. Manual data cleaning is prohibitively expensive, necessitating efficient and scalable algorithms for detecting defective samples.

Second, generative poetry models are subject to additional constraints, such as adherence to poetic meter and rhyme, alongside requirements for coherence and meaningfulness. These constraints often introduce spelling and grammatical errors in generated poems. Such errors can be eliminated using a ranker based on approaches similar to those employed for dataset cleaning.

In this paper, we address the problem of detecting linguistic defects in Russian poem texts, both in training datasets for large language models and in poems generated by these models. The unique characteristics of poetic texts, such as their stylistic and structural features, make it challenging to directly apply models and algorithms developed for other domains. This domain shift often leads to a degradation in performance, as detailed in Section \ref{sec:problem}.

Our contributions are as follows:

\begin{itemize}
    \item \textbf{The RUPOR Dataset}: A novel, human-labeled dataset for grammatical error detection, correction, and spellchecking in Russian poetry and prose. To the best of our knowledge, this is the first dataset specifically designed for the Russian poetry domain. Its size and diversity make it well-suited for evaluating grammatical error detection models and spell checkers under domain shift conditions. The dataset is described in Section~\ref{sec:data_rupor}.
    
    \item \textbf{Synthetic\_GED}: An additional dataset with rule-based distortions, designed to simulate various types of linguistic defects in natural Russian-language texts. This dataset is detailed in Section~\ref{sec:data_synthetic}.
    
    \item \textbf{Comprehensive Evaluation}: A systematic comparison of different approaches for detecting linguistic defects in Russian-language texts.
    Specifically, we demonstrate that:  
    \begin{enumerate}
        \item Perplexity is not a reliable indicator of linguistic anomalies in poetry.  
        \item Outlier detection approaches failed to achieve performance exceeding the random-guess baseline.  
        \item Supervised classification methods trained on synthetic data achieve moderate to strong detection performance. 
        \item Language models pretrained extensively on Russian-language data achieve higher classification performance than modern multilingual models, even those with larger capacities.
    \end{enumerate}
    These results are discussed in Section~\ref{sec:evaluation}.
\end{itemize}

\section{Related work}

In automatic poetry generation systems, grammaticality assessment has long been an integral part of the final evaluation of generated poems and songs.
\citet{manurung2004evolutionary} argue that the properties of well-formed poetry are poeticness, grammatically, and meaningfulness. Grammatical quality assessment can be performed by humans or by using algorithms and auxiliary language models.

Human assessment of grammaticality is described in \citet{Rstvold2020SentimentalPG, Agarwal2020AcrosticPG, ram2021say}. Human involvement in this process makes it extremely expensive, poorly scalable, and in some cases raises questions of reproducibility \citep{lau2017grammaticality}.

Automatic assessment of grammaticality comes down to either calculating the probabilistic properties of the text, in particular perplexity, or through the use of auxiliary language models finetuned for this task.

Utilization of text probability or perplexity are described in \citet{hu2024poetrydiffusion,zhang-etal-2023-lingxi,zugarini2019neural,yi2018automatic,che2017maximum,yan2016poet}.
However, the perplexity can be an unreliable basis, as noted by \citet{wang2022perplexity,kuribayashi-etal-2021-lower}.
Moreover, low perplexity may be characteristic of boring texts, which, in the case of poetry, may mean a decrease in the perceived quality and value of these texts. We will make an independent assessment of the applicability of perplexity as a criterion of grammaticality in Section \ref{sec:perplexity}.

Using a separate finetuned model to assess the grammaticality of text in generative poetry systems is less popular.
\citet{zhao2022automatic} train and then use a separate model, which they call text-fluency-checker.
For training, they use a synthetic dataset where the examples of normal texts (positive samples) are extracted from the classical Chinese poems’ dataset, and to obtain texts with defects, they apply rules that introduce distortions into the texts of positive samples.
\citet{zhu2020gruen} propose to automatically evaluate the generative poetry using a BERT-based model to evaluate the linguistic quality of the generated text based on several criteria including grammaticality.

The task of detecting and correcting grammatical errors is also widely studied outside the poetic domain. Approaches used to solve these problems include 
sentence acceptability judgments using LLM prompting~\citep{dentella2024testing, hu2023prompting},
analysis of internal text representations of encoder models~\citep{li2021bert}, topological analysis of attention maps~\citep{cherniavskii-etal-2022-acceptability}, outlier detection~\citep{xu-etal-2023-comparative}, token-level probabilistic properties of texts~\citep{gralinski2024oddballness}, and ready-to-use libraries~\citep{zhao2024pyod2}. Most grammatical error correction (GEC) approaches focus on isolated sentences; however, the cross-sentence GEC task, as studied by \citet{chollampatt-etal-2019-cross} and \citet{wang2022cctc}, is particularly relevant for detecting grammatical errors in poetry texts lacking clear sentence boundaries. 

For a number of languages, datasets and benchmarks have been created~\citep{warstadt2020blimp, Hanyue_Du_CIKM23, rothe-etal-2021-simple, juzek2024syntactic} that can be used for training and/or evaluation of GED models. Collecting and annotating data for the GED remains a labor-intensive task for annotators, so it is common practice to generate synthetic data with distortions that mimic errors in real texts \citep{martynov-etal-2024-methodology,ye2023mixedit,kiyono-etal-2019-empirical}.

\section{Problem Definition and Task Description}
\label{sec:problem}

When addressing grammatical error detection in poetic texts, it is crucial to account for unique linguistic and structural characteristics of poetry. The following outlines key challenges that must be addressed to mitigate the effects of domain shift when applying general-purpose NLP tools to poetry.

\begin{itemize}
\item Segmentation of the poem text into sentences is often impossible or hard to be performed by prose-oriented NLP tools. Line boundaries in poems do not always correspond to syntagma boundaries, for example due to the phenomenon of enjambment~\cite[Page~79]{baldick1994concise}. More natural is division into stanzas~\cite[Page~242]{baldick1994concise}, but even such structural units in rare cases are not appropriate. The ultimate solution to this problem through the processing of full poem texts may be unacceptable due to the limitations on the LM context length.

\item Neologisms, occasionalisms~\cite[Page~169]{baldick1994concise}, slang~\footnote{\url{https://genius.com/albums/Rap-genius/Rap-genius-slang-dictionary}}, active formation of compound words, the use of productive prefixes and suffixes complicate the applicability of n-gram probabilistic language models. In extreme cases~\citep{carroll2001jabberwocky}, a poem may consist almost entirely of words invented by the author and still be completely grammatical.

\item The poetic tradition pushes the boundaries of what is acceptable from a grammatical point of view. As an example, the word order in poetry differs greatly from the norm in ordinary texts. For the Russian language poetry, one can distinguish such systematic deviations as the inversion of the adjective-noun order in Noun Phrase, various non-projective dependencies, and so on. 


\item The poetic domain is characterized by a great diversity of genres and forms, with their own unique features in terms of the grammatical and lexical means used. This diversity complicates the definition of the boundaries of the norm and creates the risk of false detection of defects for texts with authorial innovations, unusual epithets or metaphors -- see \textit{zaum}~\footnote{\url{https://library.fiveable.me/key-terms/world-literature-ii/zaum}} and \textit{pirozhki}~\footnote{\url{https://pikabu.ru/story/likbez_po_mikropoyezii_pirozhki_poroshki_i_prochee_6881355}} as the examples.
\end{itemize}

Given the above, it is evident that machine learning approaches relying solely on statistical patterns derived from prose may struggle to generalize effectively to the poetry domain. This observation underscores two critical implications: (1) the inclusion of labeled poetry data in training datasets is essential, and (2) labeled poetry data is indispensable for evaluating the performance of grammatical error detection models.

It is important to note that linguistic defects in poetic texts may either be irreparable or their correction could result in undesirable degradation of poetic qualities, particularly the distortion of meter and rhyme. Consequently, the optimal strategy may involve excluding such defective texts from the training set rather than attempting to correct them. In such cases, instead of applying per-token defect annotation and correction, it is both sufficient and preferable to classify the entire text as either defective or normal.

The last thing to consider when solving this problem is the need to process a very large amount of data required to train modern language models with transformer architecture. Thus, the resource intensity or expensiveness of the approach may limit its applicability in data engineering practice.


\section{Data}
\label{sec:data}

The dataset used in this study consists of a combination of natural and synthetic samples in Russian and English, covering multiple domains, including Russian poetry and short text fragments from diverse sources. The majority of the data is in Russian, with English-language samples included to facilitate cross-evaluation of approaches, given the significant differences in morphology and syntax between the two languages.

For the Russian language, the data is sourced from publicly available datasets (described in Section~\ref{sec:data_third_party}) as well as several previously unpublished datasets (detailed in Sections~\ref{sec:data_rupor} and~\ref{sec:data_synthetic}).

The combined dataset has the following key features:

\begin{itemize}
    \item The \texttt{RUPOR} poems (Section~\ref{sec:data_rupor}) are included \textbf{only} in the test split and are not used in fine-tuning or fitting statistical models for unsupervised methods.
    \item The training split primarily consists of samples with artificial distortions (Section ~\ref{sec:data_synthetic}), supplemented by a small portion of hand-labeled data from third-party datasets (Section~\ref{sec:data_third_party}). Total number of samples in this split is 1897999.
\end{itemize}

This dataset composition ensures that all compared approaches are evaluated under strict domain shift conditions, providing a robust test of their generalizability and adaptability.

\subsection{RUPOR}
\label{sec:data_rupor}

\textbf{RUPOR} (\textbf{RU}ssian \textbf{P}oetry \textbf{O}rthographic \textbf{R}eference) is a human-labeled dataset for spell checking, grammatical error detection and correction in Russian-language poems and short prose texts. None of these samples are duplicated in other open-sourced Russian-language datasets mentioned in Section~\ref{sec:data_third_party}. The novelty and utility of the data in the RUPOR dataset are determined by the following factors:

\begin{itemize}

\item A significant difference in the lexicon and the presence of a large amount of new 2- and 3-grams compared to third-party datasets. See relevant lexicographic analysis results in Section~\ref{sec:dataset_analysis}.

\item Differences in the data format: poems in \texttt{RUPOR poetry} are presented as stanzas and groups of stanzas, while in \texttt{RUPOR prose} a lot of data is presented as several sentences. For third-party datasets, the data is usually the isolated sentences.

\item The syntax and rhetorical devices in \texttt{RUPOR poetry} are very different from those typically presented in third-party datasets.

\end{itemize}

Each sample in the \texttt{RUPOR} dataset is a pair of 1) corrupted (or linguistically unacceptable) text and 2) fixed or correct one. In some cases one of the text in a pair can be missing. Corrupted text is missing for poems that does not contain any defects. Fixed text is missing when correction is not feasible. Statistics of these sample types is shown in Table \ref{tab:rupor_info1}.

\begin{table}[h!]
\centering
\begin{tabular}{lcc}
\toprule
 & \multicolumn{2}{c}{\textbf{Num. of samples}} \\
\cmidrule(lr){2-3}   
\textbf{Sample type} & poetry & prose \\
\hline
corrupted+fixed & 5133 & 9133 \\
corrupted only & 3069 & 474 \\
correct only & 3971 & 5714 \\
\hline
\makecell[r]{\textbf{Total:}} & 12173 & 15321
\end{tabular}
\caption{The \texttt{RUPOR} dataset sample types.}
\label{tab:rupor_info1}
\end{table}

The following defects were considered by assessors when labeling the poem texts:
\begin{itemize}
\item \textit{Grammar}: Issues include incorrect grammatical forms of words, such as errors in subject-predicate agreement, adjective-noun agreement, misuse of prepositions, and improper construction of grammatical analytical forms involving auxiliary verbs, among other grammar-related problems.
\item \textit{Linguistic acceptability}: Issues include word repetition, pleonasm~\citep[Page~57]{Alexandrova2014}, grammatical gender inconsistency, and similar problems.
\item \textit{Tokenization}: Errors such as multiple words merged into one or a single word incorrectly split into multiple tokens.
\item \textit{Punctuation}: Missing or unnecessary commas, periods, or other punctuation marks.
\item \textit{Misspelling}: Incorrectly spelled words.
\end{itemize}

The token-level locations of the above defects are not marked in the samples, as this would have complicated the dataset construction process and possibly limited the quality of the marking - see the relevant arguments in \citet{sakaguchi2016reassessing}. 

The following issues were not considered defects in poems:
\begin{itemize}
  \item Illiterate speech, intentionally introduced into the text by the author. 
  \item Broken poetic meter, poor or missing rhymes.
  \item Use of hyphen symbol instead of dash.
  \item Improper lower/upper casing.
  \item Inappropriate language.
  \item Inconsistency of Cyrillic letters {\"e}~\footnote{pronounced as "yo"} and e~\footnote{pronounced as "ye"} usage.
\end{itemize}


\subsection{Synthetic Data}
\label{sec:data_synthetic}

The \textbf{synthetic\_GED} dataset comprises approximately 1 million samples and is designed for training text anomaly detection models. The creation of this large-scale synthetic dataset is motivated by the scarcity of manually labeled data, particularly in the domain of poetry. Each sample in the dataset consists of an original text (e.g. a poem stanza) and its distorted version. The distortions are generated using a combination of handcrafted rules and statistical substitutions of words and collocations. A comprehensive overview of the dataset, including its generation process, is provided in Appendix~\ref{sec:synthetic_ged_overview}.

\subsection{Third-party Datasets}
\label{sec:data_third_party}

Third-party datasets are as follows:


\texttt{RLC-Crowd} dataset~\footnote{\url{https://github.com/Russian-Learner-Corpus/rlc-crowd}}: ~10000 pairs from the Russian Learner Corpus (RLC) corrected by users of the Toloka crowd-sourcing platform. We decided not to use the second dataset from the rlc-annotated repo~\footnote{\url{https://github.com/Russian-Learner-Corpus/rlc-annotated}} due to serious quality issues found during analysis of its data.

\texttt{RuCoLa}~\footnote{\url{https://huggingface.co/datasets/RussianNLP/rucola}}: Russian Corpus of Linguistic Acceptability, ~13000 sentences labeled as linguistically acceptable or not.

\texttt{SAGE}~\footnote{\url{https://huggingface.co/datasets/ai-forever/spellcheck_punctuation_benchmark}}: four datasets from Russian Spellcheck Punctuation Benchmark~\citep{martynov2023augmentation}, each contains the pairs of defective and corrected sentences in Russian language.

Non-Russian samples come from the following sources:

Corpus of Linguistic Acceptability\footnote{\url{https://nyu-mll.github.io/CoLA/}}: 9594 English sentences labeled as acceptable/non acceptable.

\texttt{JFLEG}~\citep{napoles-etal-2017-jfleg}: an English grammatical error correction corpus containing defective sentences and corrections for them. All samples from JFLEG are included in test split.


\texttt{MultiGED-2023}~\citep{volodina-etal-2023-multiged}: training data from Shared task on Multilingual Grammatical Error Detection.

Multilingual Grammar Error Correction~\footnote{\url{https://huggingface.co/datasets/juancavallotti/multilingual-gec}} dataset: synthetic data for sentence-level grammatical error correction (GEC) in several languages. We import English-language samples only.

Several other GEC datasets for English are available but were excluded to prevent overloading the dataset. For a comprehensive overview of existing GEC datasets, we refer readers to \citet{bryant-etal-2023-grammatical}. Additionally, we excluded the BLiMP~\citep{warstadt2020blimp} and RuBLiMP~\citep{taktasheva-etal-2024-rublimp} datasets, as they contain artificially constructed corruptions, whereas our focus is on naturally occurring human errors.

\section{Evaluation of Anomaly Detection Approaches}
\label{sec:evaluation}

In this section, we describe the evaluation setups and metrics for several Text Anomaly Detection (TAD) and Grammatical Error Detection (GED) approaches. We also present an analysis of the results and discuss their practical implications. In all experiments, the model or algorithm was tasked with identifying whether a text contained at least one linguistic defect from those listed in Section~\ref{sec:problem}. We used the $F_{0.5}$ score as the evaluation metric because it prioritizes precision over recall -- a critical requirement for our task~\citep{ng-etal-2014-conll}. To ensure a balanced evaluation, the distribution of the \texttt{label=0} and \texttt{label=1} classes in the evaluation set was adjusted via undersampling so that the number of samples for each class was equal within each domain.

The $F_{0.5}$ score is calculated using the following formula:

\begin{equation}
  \label{eq:F05}
  F_{0.5} = 1.25 \frac{Prec \cdot Recall}{0.25 \cdot Prec + Recall}
\end{equation}

where $\text{Prec}$ and $\text{Recall}$ denote precision and recall, respectively. 

To establish a baseline performance, consider the case where a classifier predicts $y=1$ for all samples. In this scenario, precision $\text{Prec}=0.5$ and recall $\text{Recall}=1.0$, yielding the maximum performance score for a random-guess classifier:

\begin{equation}
  \label{eq:F05_baseline}
  F_{0.5} = 1.25 \frac{0.5 \cdot 1.0}{0.25 \cdot 0.5 + 1.0} \approx 0.56
\end{equation}

\subsection{Perplexity}
\label{sec:perplexity}

Perplexity is often used as a mean to assess the quality of generative poetry models~\citep{yi2018automatic,yan2016poet,che2017maximum,zugarini2019neural,zhang-etal-2023-lingxi,hu2024poetrydiffusion} due to the simplicity of calculation and the lack of need to somehow finetune the language model. 
However, there is evidence that perplexity is not a reliable metric for detecting defects and anomalies in text \citep{wang2022perplexity, kuribayashi-etal-2021-lower}. To investigate this further, we computed the perplexity scores using the Qwen2.5-3B model~\footnote{\url{https://huggingface.co/Qwen/Qwen2.5-3B}} for both corrupted and corrected texts. The results, presented in Table \ref{tab:ppl_delta}, reveal significant variability in perplexity scores, which limits the utility of absolute perplexity as a criterion for distinguishing between correct and defective texts, even within a single domain. A visual illustration of this issue in the context of poetry is provided in Appendix~\ref{app:perplexity_poems}.

\begin{table*}[h!]
\centering
\begin{tabular}{lcccccccc}
\toprule
  & \multicolumn{2}{c}{\textbf{corrupted texts}} & \multicolumn{2}{c}{\textbf{fixed texts}} & \multicolumn{2}{c}{\textbf{share of samples}} \\
\cmidrule(lr){2-3} \cmidrule(lr){4-5} \cmidrule(lr){6-7}  
\textbf{Domain} & \(\overline{ppl}\) & $\sigma$ & \(\overline{ppl}\) & $\sigma$ & \(\Delta_{ppl}\)>0 & \(\Delta_{ppl}\)<0 & \makecell[c]{KS test \\ p-value} \\
\hline

RUPOR poetry & 59.5 & 42.6 & 49.5 & 33.3 & 0.85 & 0.12 & 4.4e-37 \\
RUPOR prose & 982.9 & 8103.2 & 795.5 & 21234.4 & 0.82 & \textbf{0.16} & 3.1e-33 \\
synthetic\_GED & 862.6 & 9955.8 & 617.4 & 12210.3 & 0.96 & 0.03 & 0.0e+00 \\
\hline
rlc-toloka (ru) & 227.9 & 608.3 & 119.6 & 409.8 & 0.91 & 0.07 & 3.9e-178 \\
SAGE RUSpellRU & 334.5 & 1376.8 & 91.8 & 209.5 & 0.96 & 0.04 & 1.9e-164 \\
SAGE MultidomainGold & 366.4 & 5081.6 & 149.7 & 923.8 & 0.88 & 0.1 & 7.5e-39 \\
\hline
JFLEG (en) & 903.2 & 3817.9 & 432.8 & 1631.2 & 0.85 & 0.13 & 1.8e-92 \\

\bottomrule
\end{tabular}
\caption{Differences in perplexity scores between correct and corrupted texts. \(\Delta_{ppl} = ppl_{corrupted} - ppl_{fixed} \), \(ppl\) stands for perplexity, $\sigma$ is standard deviation, kstest is Kolmogorov-Smirnov test for \(\{ppl_{corrupted}\}\) and \(\{ppl_{fixed}\}\).}
\label{tab:ppl_delta}
\end{table*}

The obtained p-values of the Kolmogorov-Smirnov test (the column captioned \textit{ks test p-value}) show that the null hypothesis about the coincidence of perplexity distributions for damaged and corrected texts can be rejected with a high level of significance. Thus, LM distinguishes these texts well.

\begin{figure}[h!]
\includegraphics[width=0.5\textwidth]{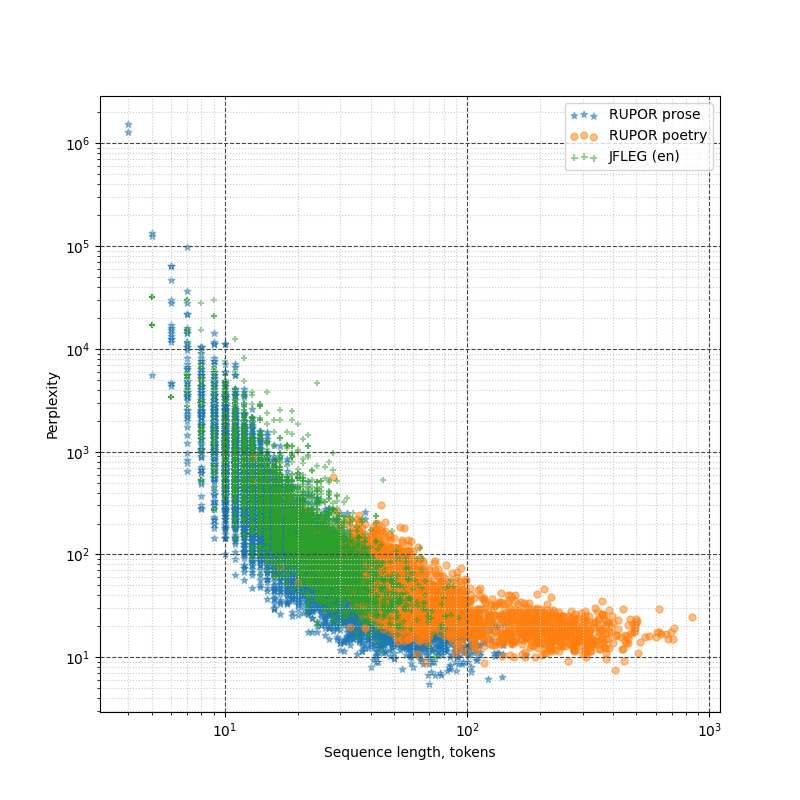}
\caption{Relationship between perplexity and token sequence length, illustrating how perplexity varies as the sequence length increases.}
\label{fig:numtoken_perplexity_correlation}
\end{figure}

The arithmetic mean of perplexity for corrupted texts (\(ppl_{corrupted}\)) is greater than for corrected texts (\(ppl_{fixed}\)) among almost all datasets. This result is expected and is fully consistent with the definition of perplexity. But there are a significant number of samples (up to 13\% in case of \texttt{RUPOR poetry}) when perplexity  \textbf{increases} after correcting defects, that is, \(\Delta_{ppl}\)<0. One of the reasons for this result may be an influence of the token sequence length on the calculated perplexity, especially in the case of relatively short texts. The Figure \ref{fig:numtoken_perplexity_correlation} visually confirms that such an influence does take place. The corresponding values of the Pearson's correlation coefficient are given in the Table \ref{tab:tbl_ppl_numtokens_pearson}. They show that the correlation is weak, but should not be ignored.

\begin{table}[h!]
\centering
\begin{tabular}{lcc}
\toprule
Domain & $\rho$ & p-value \\
\hline
RUPOR poetry & -0.36 & 1.4e-150 \\
RUPOR prose & -0.034 & 0.0013 \\
JFLEG (en) & -0.244 & 2.3e-70 \\
\bottomrule
\end{tabular}
\caption{Person's correlation coefficient $\rho$ between perplexity and token sequence length.}
\label{tab:tbl_ppl_numtokens_pearson}
\end{table}

To validate this hypothesis further, we separately calculated \(\Delta_{ppl}\) for cases where 1) the length of the defective and corrected texts is the same, 2) the corrected text is longer, 3) the corrected text is shorter. In all cases, the length was calculated as the number of tokens returned by the tokenizer of the LM used. The results are shown in Table \ref{tab:ppl_delta_wro_numtokens}. It is evident that an abnormal change in perplexity during text correction is observed if correction increases the length of the text: see the column \(\delta_{numtok}\)>0 , \(\Delta_{ppl}\)>0 .

\begin{table*}[h!]
\centering
\begin{tabular}{lcccccc}
\toprule
 & \multicolumn{6}{c}{\textbf{Share of samples}} \\
\cmidrule(lr){2-7}  
 & \multicolumn{2}{c}{\(\delta_{numtok}\)=0} & \multicolumn{2}{c}{\(\delta_{numtok}\)>0} & \multicolumn{2}{c}{\(\delta_{numtok}\)<0} \\
\cmidrule(lr){2-3}   \cmidrule(lr){4-5}   \cmidrule(lr){6-7} 
\textbf{Domain} & \(\Delta_{ppl}\)>0 & \(\Delta_{ppl}\)<0 & \(\Delta_{ppl}\)>0 & \(\Delta_{ppl}\)<0 & \(\Delta_{ppl}\)>0 & \(\Delta_{ppl}\)<0 \\

\hline

RUPOR poetry & 0.3 & 0.02 & 0.29 & 0.1 & 0.26 & 0.0 \\
RUPOR prose & 0.35 & 0.02 & 0.22 & \textbf{0.14} & 0.25 & 0.0 \\
synthetic\_GED & 0.12 & 0.0 & 0.41 & 0.03 & 0.44 & 0.0 \\
\hline
rlc-toloka (ru) & 0.34 & 0.01 & 0.3 & 0.06 & 0.27 & 0.0 \\
SAGE RUSpellRU & 0.13 & 0.01 & 0.15 & 0.03 & 0.67 & 0.0 \\
SAGE MultidomainGold & 0.22 & 0.03 & 0.27 & 0.06 & 0.39 & 0.02 \\
\hline
JFLEG (en) & 0.2 & 0.02 & 0.35 & 0.11 & 0.3 & 0.0 \\

\bottomrule
\end{tabular}
\caption{Perplexity change \(\Delta_{ppl}\) with respect to text length difference \(\delta_{numtok}\). \(\Delta_{ppl} = ppl_{corrupted} - ppl_{fixed} \) , \(\delta_{numtok}\) = numtokens(fixed) - numtokens(corrupted) }
\label{tab:ppl_delta_wro_numtokens}
\end{table*}

All these results above can be summarized as follows. Using LM perplexity to detect poems with linguistic anomalies should be done with great caution taking into account, in particular, the factor of text length.

In addition to perplexity, several token-level metrics can be derived from the distribution of LLM token logits, such as token probability, token rank, and entropy. These metrics have been used to identify anomalies in text \citep{cha2024pre,gralinski2024oddballness}. However, as shown in Appendix~\ref{app:token-level-anomaly}, their effectiveness in detecting defects is limited, at least for Russian-language poetry.

\subsection{Outliers Detection}
\label{sec:outliers_detection}

The outlier detection methods are based on the assumption that we have a corpus of normal texts for which statistical model is constructed. Detecting anomalies involves evaluating how much a given text deviates from the permissible variations derived from this model. 

For this task, we utilized the \texttt{RuRoberta-large} encoder model~\footnote{\url{https://huggingface.co/ai-forever/ruRoberta-large}} (355M parameters) to generate 1024-dimensional text embeddings. The embedding vector was computed by averaging the contextual embeddings of the text tokens, excluding the special tokens \texttt{<cls>} and \texttt{</s>}. We employed outlier detection algorithms from the \texttt{PYOD} library~\citep{zhao2024pyod2} to classify texts as inliers (\(y = 0\)) or an outlier (\(y = 1\)) based on their embedding vectors. All \texttt{PYOD} models were trained on 10,000 Russian-language samples from the training split described in Section~\ref{sec:data} and evaluated on 10,000 samples from the \texttt{RUPOR} dataset. Given the ground truth labels indicating whether a text is corrupted or correct, we compute the \(F_{0.5}\) score. The evaluation results presented in Table~\ref{tab:pyod-results-best-algorihms} show that no anomaly detection algorithm in this setting was able to surpass the random classification bound.

\begin{table}[h!]
\centering
\begin{tabular}{lcc}
\toprule
\textbf{Algorithm} & $F_{0.5}$ \\
\hline

KPCA                                       & 0.56  \\
Kernel Density Functions (KDE)             & 0.56  \\
Deep Isolation Forest (DIF)                & 0.50  \\
Angle-based Outlier Detector (ABOD)        & 0.47  \\
One-class SVM (OCSVM)                      & 0.44  \\

\bottomrule
\end{tabular}
\caption{Results of evaluating \texttt{PYOD} outlier detection algorithms on the \texttt{RUPOR} poetry samples using RuRoberta-large embeddings. Top-5 scores are shown.}
\label{tab:pyod-results-best-algorihms}
\end{table}

\subsection{Zero-shot Detection of Linguistic Anomalies with LLMs}
\label{sec:zeroshot_llm_prompting}

Leveraging a variety of high-quality, instruction-tuned open LLMs, we evaluate their effectiveness in detecting grammatical anomalies in a zero-shot prompting~\citep{brown2020language} setting.
For the experiment, we took models with a capacity from 3B to 20B.
Limitations on computing resources did not allow us to test the performance for more capacious models.

The models are given a prompt (translated from Russian) of the following form:

\textit{Analyze the grammar and spelling of the text below.
    If the text contains any grammatical or spelling errors, list them.
    Otherwise, return "no errors."}

The text being checked is added to this prompt, the result is substituted into the chat template of the instructional model.

Additionally, we evaluated Yandex Speller~\footnote{\url{https://yandex.ru/dev/speller}} using a Python wrapper~\footnote{\url{https://github.com/oriontvv/pyaspeller}}: if the web service suggests corrections, the text is considered anomalous. Due to limitations on the allowed number of requests per day for Yandex Speller API, we limited ourselves to testing all models and APIs on 1000 samples per domain. To estimate confidence intervals for all models at a 95\% confidence level, we employed the bootstrap resampling with 1000 repetitions.

The evaluation results in Table~\ref{tab:zeroshot_llm_results} demonstrate the poor performance of this approach, which may be attributed to suboptimal prompt design. However, even for very large language models, comparable limitations have been reported in similar studies~\citep{dentella2024testing}, highlighting inherent challenges in zero-shot anomaly detection tasks. The results for other test data are available in Appendix~\ref{app:zero-shot-results}.

\begin{table}[h!]
\centering
\begin{tabular}{lcc}
\toprule
 & \multicolumn{2}{c}{\texttt{RUPOR} poetry} \\
\cmidrule(lr){2-3}  
\textbf{Model/Service} & $\overline{F_{0.5}}$ & 95\% CI \\
\hline
YandexSpeller & \textbf{0.77} & 0.73, 0.80 \\
GigaChat-20B-instruct~\footnote{\url{https://huggingface.co/ai-sage/GigaChat-20B-A3B-instruct-v1.5-bf16}} & 0.57 & 0.52, 0.61 \\
T-lite-it-1.0~\footnote{\url{https://huggingface.co/t-tech/T-lite-it-1.0}} & 0.58 & 0.55, 0.61 \\
Qwen2.5-3B-Instruct~\footnote{\url{https://huggingface.co/Qwen/Qwen2.5-3B-Instruct}} & 0.39 & 0.33, 0.45 \\
Mistral-7B-Instruct~\footnote{\url{https://huggingface.co/mistralai/Mistral-7B-Instruct-v0.3}} & 0.55 & 0.52, 0.58 \\
\bottomrule
\end{tabular}
\caption{Mean $F_{0.5}$ scores, confidence interval (CI) lower and upper bounds for spellcheckers and zero-shot instruction-tuned LLMs on corrupted poem detection. CIs were calculated using bootstrap resampling with 1,000 iterations.}
\label{tab:zeroshot_llm_results}
\end{table}

\subsection{LLM-Based Supervised Text-Level Detection of Linguistic Anomalies}
\label{sec:classifier}

Detecting linguistically anomalous text can be formulated as a binary classification task. To address this, we fine-tuned several pre-trained language models on the training split of the combined dataset described in Section~\ref{sec:data} and evaluated their performance on the \texttt{RUPOR} and third-party datasets.

We selected the following transformer models pre-trained on Russian corpora:
\begin{itemize}
\item FRED-T5-1.7B~\footnote{\url{https://huggingface.co/ai-forever/FRED-T5-1.7B}} (an encoder-decoder architecture),
\item ruRoberta-large (an encoder-only architecture), and
\item rugpt3medium~\footnote{\url{https://huggingface.co/ai-forever/rugpt3medium_based_on_gpt2}} (a decoder-only architecture).
\end{itemize}
Additionally, we included the Qwen2.5-3B model as a representative of modern multilingual decoder models. Due to computational resource constraints, we limited our experiments to models with 3 billion parameters.

For fine-tuning, we employed both full supervised fine-tuning and parameter-efficient LoRa  techniques. To ensure the robustness of our results, each foundation model was fine-tuned and evaluated three times with shuffled training data. This allowed us to compute 95\% confidence intervals for the $F_{0.5}$ scores. The evaluation metrics are summarized in Table~\ref{tab:supervised_classifiers}.

A comprehensive description of the experimental setup, along with the complete evaluation results, is provided in Appendix~\ref{app:binary-classifiers}.

\begin{table}[h!]
\centering
\begin{tabular}{lcc}
\toprule
 & \multicolumn{2}{c}{\texttt{RUPOR} poetry} \\
\cmidrule(lr){2-3}  
\textbf{Foundation LM} & $\overline{F_{0.5}}$ & 95\% CI \\
\hline
FRED-T5-1.7B & \textbf{0.863} & 0.857, 0.870 \\ 
ruRoberta-large & 0.802 & 0.787, 0.818 \\
Qwen2.5-3B & 0.810 & 0.804, 0.816 \\
Qwen2.5-3B LoRa & 0.749 & 0.710, 0.789 \\
rugpt3medium & 0.615 & 0.535, 0.694 \\
\bottomrule
\end{tabular}
\caption{Evaluation results for supervised binary classifiers on the \texttt{RUPOR} corrupted poem detection. Confidence intervals (CI) are calculated at a 95\% confidence level with a t-value of 4.3.}
\label{tab:supervised_classifiers}
\end{table}

Based on these results, supervised binary classifiers represent the only practical solution for the task of grammatical error detection in Russian-language texts.

\section{Conclusion}
\label{sec:conclusion}

In this paper, we introduced the \texttt{RUPOR} dataset, a novel resource comprising over 12,000 labeled fragments of contemporary Russian poetry, designed for detecting various forms of syntactic and lexical anomalies. Using this dataset, we evaluated the performance of multiple models and algorithms in identifying linguistic irregularities. Our results highlight the dataset's utility for research in both Russian-language generative poetry and broader linguistic error detection tasks.

\section*{Limitations}

This study has several limitations, which we outline below.

\vspace{0.5em} \noindent\textbf{Dataset Collection and Labeling Challenges}: Collecting and annotating datasets for spellchecking, grammatical error detection, and correction tasks is highly labor-intensive. The complexity of identifying diverse defect types—such as punctuation, spelling, grammar, and pragmatics—may influence the distribution of defect examples in the dataset and the accuracy of their labeling. The extent of this impact warrants further investigation in future work.

\vspace{0.5em} \noindent\textbf{Rare anomalies}: The frequencies of different types of linguistic defects in real texts vary greatly. Some defect types may be underrepresented in the model evaluation due to this implicit imbalance.

\vspace{0.5em} \noindent\textbf{Potential Biases in Data Sources}: The defective texts in the \texttt{RUPOR} dataset were collected from various sources, which may introduce biases. These biases could affect the assessment of generalization capabilities of models. For example, the mistakes made by foreigners learning Russian may systematically differ from the mistakes made by native speakers in different contexts.

\vspace{0.5em} \noindent\textbf{Extrapolation to larger LLMs}: Due to resource constraints, we were unable to perform experiments with language models larger than 7 billion parameters. However, the observed increase in performance metrics with larger model capacities among the tested models suggests that further improvements could be achieved by utilizing larger-scale models.

\vspace{0.5em} \noindent\textbf{Synthetic dataset quality}: A limitation of the supervised approach is its reliance on synthetic datasets for training. The effectiveness of the evaluation is inherently tied to how accurately the synthetic text distortions replicate real-world defects, both within the evaluation dataset and in unseen data. To address this, future work should include systematic ablation studies to identify the optimal share of synthetic samples generated by different rules and edit statistic based algorithms, ensuring better generalization to real-world scenarios.

\vspace{0.5em}
\noindent\textbf{Limitations of Proprietary Spellcheckers and Chatbots}: 
In this study, we did not evaluate the performance of proprietary spellcheckers and chatbots in detecting defective text using zero-shot prompting via their APIs except for YandexSpeller. This omission is primarily due to the significant effort required to carefully design and optimize prompt formats for each individual system, which is both labor-intensive and time-consuming. Despite this limitation, we acknowledge that such an evaluation could yield valuable insights, given the remarkable progress these systems have demonstrated in a wide range of text processing tasks. Future work could address this gap by systematically exploring the capabilities of proprietary systems in defective text detection.

\section*{Ethical Consideration}
\label{sec:ethical}

\vspace{0.5em} \noindent\textbf{Content Sensitivity}: We did not filter the collected samples for offensive content, which may include text that could offend religious, political, or ethical sensibilities. This limitation highlights the need for careful content moderation in practical applications.

\vspace{0.5em} 
\noindent\textbf{False Positives:} A grammatical anomaly detection model that produces false positives may systematically bias the composition of a dataset when used for cleaning training data. Specifically, certain genres or topics could be disproportionately excluded due to erroneous classification as grammatically anomalous. Similarly, in generative poetry systems, false positives may adversely affect the ranking of generated texts, leading to the prioritization of less creative or linguistically diverse outputs. This underscores the importance of minimizing false positives to ensure fair and representative data curation and evaluation.



\vspace{0.5em} \noindent\textbf{AI-assistants Help.} This paper was proofread and improved using the DeepSeek assistant to correct grammatical, spelling, and stylistic errors, as well as to enhance readability. As a result, certain portions of the text may be flagged as AI-generated, AI-edited, or human-AI co-authored by detection tools. However, all ideas, research, and contributions remain entirely our own.




\bibliography{custom}

\appendix

\section{Appendix}
\label{sec:appendix}

\subsection{Dataset and Supplementary Code Repository}
\label{app:repository}

The repository \href{https://github.com/anonymous\_repository}{GitHub} contains the following datasets and code available under the MIT license:

\begin{itemize}

\item \textbf{Datasets:}
\begin{itemize}
\item The \texttt{RUPOR} dataset (Section~\ref{sec:data_rupor}).
\item The \texttt{synthetic\_GED} dataset (Section~\ref{sec:data_synthetic}), including train and test splits.
\end{itemize}

\item \textbf{Code:}
\begin{itemize}
    \item Training and evaluation scripts for:
    \begin{itemize}
        \item Outlier detection experiments.
        \item Zero-shot LLM prompting.
        \item Fine-tuning binary classifiers.
    \end{itemize}
    \item Lexicographic analysis tools for dataset exploration.
    \item Code for perplexity-based experiments.
    \item Adapted implementation from \citet{cherniavskii-etal-2022-acceptability} for evaluating the topological approach using attention maps (not included in the main results).
\end{itemize}

\item All evaluation results with detailed logs.
\end{itemize}

Additionally, the top-performing model for Russian language ungrammatical text detection, as detailed in Section~\ref{sec:classifier}, is hosted at \href{https://huggingface.co/anonymous\_repository}{Hugging Face}. This model is available for public use and experimentation.





\subsection{Overview of the synthetic\_GED Dataset Generation}
\label{sec:synthetic_ged_overview}

The \texttt{synthetic\_GED} dataset is generated using a multi-step algorithm that combines grammatically correct phrases and sentences from diverse sources with systematic distortions. Below, we describe the sources of correct text and the rules used to introduce grammatical errors and anomalies, simulating those found in natural texts.

\subsubsection{Sources of Correct Text}
The following sources were used to collect linguistically correct phrases and sentences:

\begin{itemize}
    \item \textbf{Wikipedia}: Titles of movies, books, and other relevant content.
    \item \textbf{Syntactic Analyzer Output}: Verb phrases (VP) and noun phrases (NP) extracted using a rule-based syntactic analyzer~\footnote{\url{https://github.com/anonymous-repository}}.
    \item \textbf{LLM-Generated Texts}: Short texts generated by a large language model (LLM) using keyword-based prompts to maximize lexical coverage and provide diverse usage examples.
    \item \textbf{RuSimpleSentEval}: Sentences from the RuSimpleSentEval dataset~\footnote{\url{https://github.com/dialogue-evaluation/RuSimpleSentEval}}.
    \item \textbf{Phraseological Expressions}: Idioms and fixed expressions sourced from Wiktionary~\footnote{\url{https://ru.wiktionary.org/}} and the website~\footnote{\url{https://moiposlovicy.ru/frazeologizmy}}.
    \item \textbf{Poetry Fragments}: Excerpts of 1–2 stanzas from poetic works.
    \item \textbf{Quasipoetry}: Prose paragraphs reformatted as poetry, with 4 to 10 words per line and the first letter of each line capitalized.
\end{itemize}

\subsubsection{Rules for Introducing Distortions}
The following rules and algorithms were applied to introduce grammatical errors and anomalies into the correct texts:

\begin{itemize}
    \item \textbf{Grammatical Form Distortion}: Altering the grammatical form of words involved in syntactic dependencies, such as disrupting subject-verb agreement in Russian.
    \item \textbf{Preposition Errors}: Replacing or deleting prepositions, which often results in syntax violations due to case requirements in Russian.
    \item \textbf{Spelling Errors}: Injecting typos and spelling errors collected from logs of a rule- and dictionary-based spell checker developed for a generative Russian poetry project.
    \item \textbf{Word Splitting and Merging}: Randomly splitting or merging words, including detaching prefixes from verbs and adverbs.
    \item \textbf{Punctuation Errors}: Randomly inserting or removing commas.
    \item \textbf{Augmentex Library}: Using the Augmentex library~\footnote{\url{https://github.com/ai-forever/augmentex}} to inject misspellings based on rule-based and statistical approaches.
\end{itemize}

The rules for grammatical form distortion and preposition errors rely on part-of-speech analysis and syntactic parsing performed using the \texttt{ufal.udpipe} library~\footnote{\url{https://github.com/ufal/udpipe}} (v1.3.0.1) with the \texttt{SynTagRus} model.

\subsection{Dataset Analysis}
\label{sec:dataset_analysis}

To evaluate the novelty and linguistic characteristics of the \textbf{RUPOR} (Section~\ref{sec:data_rupor}) and \textbf{Synthetic\_GED} (Section~\ref{sec:data_synthetic}) datasets, we conducted a lexicographic analysis. This analysis was designed to compare their vocabulary and N-gram distributions with those of the third-party datasets discussed in Section~\ref{sec:data_third_party}.

We focused on vocabulary size and N-gram statistics, excluding punctuation and numerical tokens, to ensure a fair comparison. The results of this analysis are summarized in Table~\ref{tab:vocabulary}, which highlights key differences and similarities across the datasets.

\begin{table*}[h!]
\centering
\begin{tabular}{lcccc}
\toprule
Dataset           & Num. of words & Unique words & Unique 2-grams & Unique 3-grams \\
\hline
RUPOR poetry         & 361991        & 58250        & 191026         & 227227     \\    
RUPOR prose          & 216422        & 41898        & 109299         & 121090     \\    
synthetic\_GED        & 23778033      & 843268       & 7449623        & 11434652  \\     
\hline
SAGE MultidomainGold & 401547        & 44623        & 161574         & 215474     \\    
SAGE RUSpellRU       & 90679         & 19326        & 44990          & 50303      \\    
\bottomrule
\end{tabular}
\caption{Vocabulary and n-gram statistics of the \texttt{RUPOR} and \texttt{Synthetic\_GED} datasets, as well as some third-party datasets.}
\label{tab:vocabulary}
\end{table*}

Both the \texttt{RUPOR} and \texttt{Synthetic\_GED} datasets substantially increase the lexical diversity of data by adding samples with new words as shown in Tables \ref{tab:vocabulary_novelty} and \ref{tab:vocabulary_overlap}.

\begin{table*}[h!]
\centering
\begin{tabular}{lccc}
\toprule
Third-party dataset & RUPOR poetry & RUPOR prose & synthetic\_GED \\
\hline
rlc-toloka (ru)                        & 31177        & 21634       & 722512        \\
SAGE RUSpellRU                         & 32752        & 23784       & 729245        \\
SAGE MultidomainGold                   & 29560        & 19644       & 713333        \\
SAGE MedSpellchecker                   & 37654        & 29188       & 740868        \\
SAGE GitHubTypoCorpusRu                & 36536        & 27588       & 738156        \\
\bottomrule
\end{tabular}
\caption{Number of new unique words in the \texttt{RUPOR} and \texttt{Synthetic\_GED} datasets that are not in third-party datasets.}
\label{tab:vocabulary_novelty}
\end{table*}

\begin{table*}[h!]
\centering
\begin{tabular}{lccc}
\toprule
Third-party dataset & RUPOR poetry & RUPOR prose & synthetic\_GED \\
\hline
rlc-toloka (ru)                        & 0.119        & 0.169       & 0.026         \\
SAGE RUSpellRU                         & 0.108        & 0.15        & 0.017         \\
SAGE MultidomainGold                   & 0.121        & 0.168       & 0.038         \\
SAGE MedSpellchecker                   & 0.01         & 0.02        & 0.002         \\
SAGE GitHubTypoCorpusRu                & 0.036        & 0.066       & 0.006         \\
\bottomrule
\end{tabular}
\caption{Vocabulary overlap.}
\label{tab:vocabulary_overlap}
\end{table*}

An important property of the \texttt{RUPOR} dataset is that a significant proportion of samples are pairs of defective and corrected texts. The difference between the texts in each such pair represents a certain set of editing operations. The diversity of these edits is an important metric when using datasets as a benchmark as lexicographic diversity. Below we describe the methodology to qualitatively and quantitatively assess such diversity. 

\begin{figure}[h!]
\includegraphics[width=0.5\textwidth]{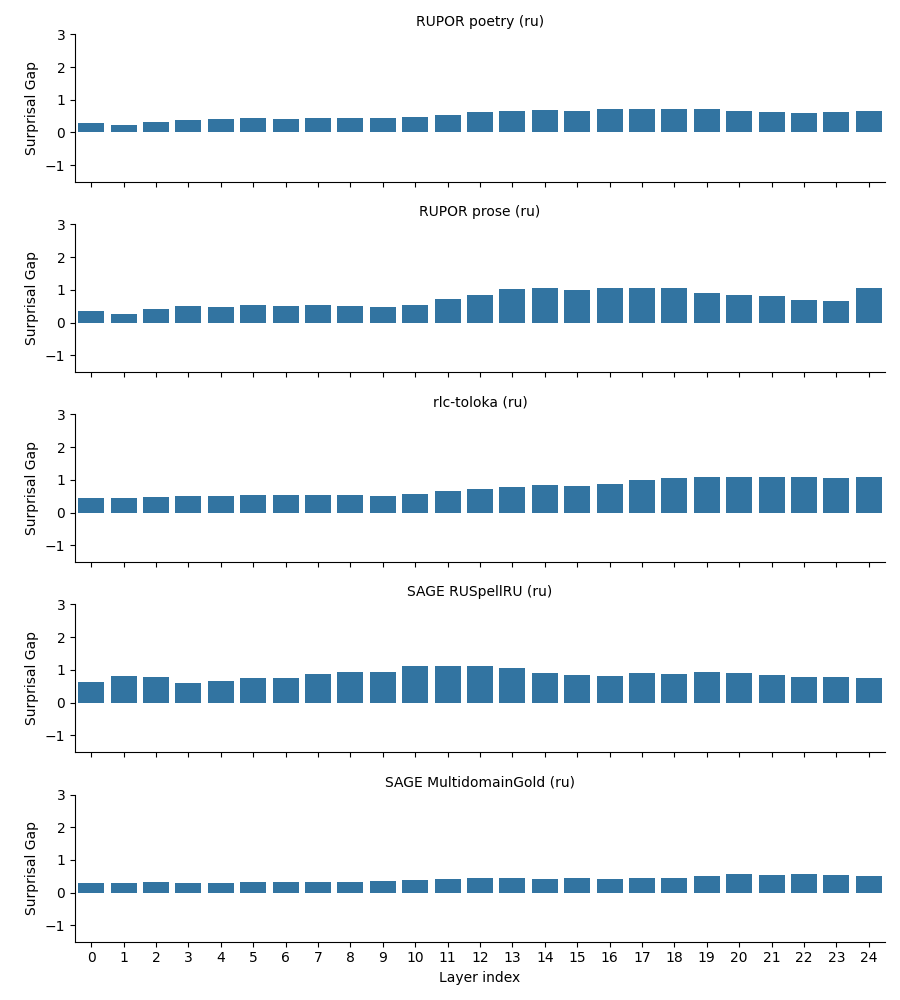}
\caption{Surprisal gaps for GMMs trained on ruRoberta-large embeddings and Russian-language text pairs.}
\label{fig:surprisal_gaps}
\end{figure}

First, we reproduced the surprisal gap calculations from \citet{li2021bert} to evaluate 1) how the approach described there would behave on Russian poetry texts and 2) to qualitatively assess the diversity of edits in the \texttt{RUPOR} dataset. The Gaussian mixture models (GMM) are trained on 2000 correct texts from combined training set, and the surprisal gaps are calculated on 200 pairs from the \texttt{RUPOR} poetry and prose, and rlc-toloka. The results are shown in Figure~\ref{fig:surprisal_gaps}.

Comparing with the calculation results in the original paper (see Figure 4 in \citet{li2021bert}), we can state that the gap values observed in our experiment are in good agreement in result for "semantic" task pairs in the original paper. Interestingly, unlike the results in the original paper, for ruRoberta-large the gap value does not show a strong drop for the last layer.

To quantitatively analyze edits in the \texttt{RUPOR} dataset, we used the \texttt{difflib} library~\footnote{\url{https://docs.python.org/3/library/difflib.html}} to build a list of word-level edits to defective text that are necessary to obtain correct text. Statistics of word-level edit categories are shown in Table \ref{tab:rucola2_edit_categories}. The edit categories are as follows:

\textbf{Spelling} category accumulates the replacements of misspelled words.

\textbf{Punctuation} category is for deletions, insertions, and replacement of punctuation symbols.

\textbf{Tokenization} category is for cases: 1) a word is split into two or more words, 2) two or more words are merged into one word.

\textbf{Other} category captures all edits not mentioned in previous 3 categories including grammar fixes.

\begin{table}[t]
\centering
\begin{tabular}{lcccc}
\toprule
         & \multicolumn{2}{c}{poetry}  & \multicolumn{2}{c}{prose} \\
\cmidrule(lr){2-3}   \cmidrule(lr){4-5}
Category & Count & Share & Count & Share \\
\hline
spelling     & 368  & 0.02 & 229 & 0.02 \\
tokenization & 1354 & 0.08 & 1852 & 0.17 \\
punctuation  & 6299 & 0.39 & 1430 & 0.13 \\
other        & 8301 & 0.51 & 7709 & 0.69 \\
\bottomrule
\end{tabular}
\caption{Word-level edit categories in the \texttt{RUPOR} dataset. \textit{Other} category includes grammar issues and other defect types.}
\label{tab:rucola2_edit_categories}
\end{table}

The most frequent number of edits required to completely correct a defective text is 1 (see Table \ref{tab:rucola2_poetry_numedits} for poetry and Table \ref{tab:rucola2_prose_numedits} for prose). However, there are also samples with a bigger number of editable defects. "Number of edits=0" stands for insertion and deletion of space around a hyphen since technically in this case the editing does not affect the words.

\begin{table}[t]
\centering
\begin{tabular}{ccc}
\toprule
 Num. of edits & Num. of samples & Share \\
\hline
0               & 4                 & 0.001 \\
1               & 2506              & \textbf{0.488} \\
2               & 939               & 0.183 \\
3               & 541               & 0.105 \\
4               & 325               & 0.063 \\
5               & 171               & 0.033 \\
>5              & 647               & 0.126 \\
\bottomrule
\end{tabular}
\caption{Number of word-level edits required to obtain a correct version of the poem in the \texttt{RUPOR} dataset.}
\label{tab:rucola2_poetry_numedits}
\end{table}

\begin{table}[t]
\centering
\begin{tabular}{ccc}
\toprule
Num. of edits & Num. of samples & Share \\
\hline
0               & 18                & 0.002  \\
1               & 7731              & \textbf{0.846} \\
2               & 969               & 0.106 \\
3               & 247               & 0.027 \\
4               & 106               & 0.012 \\
5               & 26                & 0.003 \\
>5              & 36                & 0.004 \\
\bottomrule
\end{tabular}
\caption{Number of word-level edits required to obtain a correct version of the text in prose division of the \texttt{RUPOR} dataset.}
\label{tab:rucola2_prose_numedits}
\end{table}

As noted in Section \ref{sec:problem}, poem texts are not always divisible into sentences. Therefore, in the poerty section of the dataset, the texts are either individual stanzas or entire poems consisting of 2 or more stanzas. Statistics on the number of non-empty lines (i.e., we do not include the empty line between stanzas in the count) in poems for top-10 cases are presented in Table \ref{tab:rupor_poetry_numlines}.

\begin{table}[h!]
\centering
\begin{tabular}{lcc}
\toprule

 Num. of lines & Num. of samples & Share \\
\hline
4           & 6766          & \textbf{0.825} \\
8           & 383           & 0.047  \\
3           & 213           & 0.026  \\
6           & 128           & 0.016  \\
16          & 107           & 0.013  \\
5           & 88            & 0.011  \\
20          & 84            & 0.01   \\
12          & 80            & 0.01   \\
24          & 65            & 0.008  \\
14          & 45            & 0.005  \\
\bottomrule
\end{tabular}
\caption{Statistics of lines per poem in the \texttt{RUPOR} poems.}
\label{tab:rupor_poetry_numlines}
\end{table}

Following \citet{ye2023mixedit} we estimate how well the synthetic data reproduce distortions in real texts by calculating Kullback–Leibler divergence of word-level edit frequencies.

\begin{equation}
\label{eq:kld}
 D_{KL}(P || Q) = \sum_{i} p_i \, log \frac{p_i}{q_i}
\end{equation}

where $P$ and $Q$ are frequencies of word-level edits in two datasets.

KL divergence was calculated using \texttt{scipy.stats.entropy}~\footnote{\url{https://docs.scipy.org/doc/scipy/reference/generated/scipy.stats.entropy.html}}. The results are presented in Table~\ref{tab:kl-divergence-datasets}. 

\begin{table*}[h!]
\centering
\begin{tabular}{llc}
\toprule
 Reference dataset, $P$ & Compared dataset, $Q$ & \( D_{KL}(P || Q) \) \\
\hline
 RUPOR poetry         & synthetic\_GED        & 6.06  \\
 SAGE RUSpellRU       & SAGE MultidomainGold & 7.65  \\
 RUPOR prose          & synthetic\_GED        & 8.12  \\
 SAGE MultidomainGold & SAGE RUSpellRU       & 8.83  \\
 rlc-toloka (ru)      & RUPOR prose          & 15.08 \\

\bottomrule
\end{tabular}
\caption{Minimum Kullback–Leibler divergence of edit frequencies}
\label{tab:kl-divergence-datasets}
\end{table*}


\subsection{Perplexity and Text Anomalies in Poems}
\label{app:perplexity_poems}


Perplexity, while a useful metric for evaluating language models, is not a robust indicator of linguistic fluency or grammatical correctness. This limitation can be demonstrated by comparing the perplexity scores of linguistically impeccable texts with those of distorted or unconventional variants. To illustrate this, we evaluate the perplexity of several texts using the Qwen3.5-3B language model: the first quatrain of the well-known lullaby "Twinkle, Twinkle, Little Star," a slightly distorted version of the same quatrain, and two parodies. The results, shown in Table~\ref{tab:parody_perplexities}, highlight the challenges of using perplexity as a standalone metric.

\begin{table*}[h!]
\centering
\begin{tabular}{lcr}
\toprule
\textbf{Quatrain} & \textbf{Note} & \textbf{Perplexity} \\
\hline
\makecell[l]{Twinkle, twinkle, little star, \\
How I wonder what you are! \\
Up above the world so high, \\
Like a diamond in the sky.} & original lullaby text & 5.0 \\
\hline
\makecell[l]{Twinkle, twinkle, little star, \\
How \textbf{It} wonder what you are! \\
Up above the world so high, \\
Like a diamond in the sky.} & distorted version of original & 14.5 \\
\hline
\makecell[l]{Twinkle, twinkle, little bat!\\
How I wonder what you’re at!\\
Up above the world you fly,\\
Like a teatray in the sky.} & a parody by Lewis Carroll's & 12.2 \\
\hline
\makecell[l]{Twinkle, twinkle, little star,\\
I wanna hit you with a car,\\
Throw you off a bridge so high,\\
Hope you break your neck and die.} & less popular parody & 28.8 \\
\bottomrule
\end{tabular}
\caption{Qwen3.5-3B perplexity scores for "Twinkle, twinkle ...", its parodies, and distorted version.}
\label{tab:parody_perplexities}
\end{table*}

As shown in Table~\ref{tab:parody_perplexities}, the perplexity of the distorted version (14.5) is closer to that of the first parody (12.2) than to the original text (5.0). This suggests that perplexity alone cannot reliably distinguish between grammatical errors and unconventional but valid linguistic constructs. Furthermore, the second parody, which is less conventional and more semantically divergent, has a significantly higher perplexity (28.8). These findings align with the observations of \citet{lau2017grammaticality}.


\subsection{Token-level Indicators of Linguistic Anomalies}
\label{app:token-level-anomaly}


Given a foundation language model capable of providing a probability distribution over the entire vocabulary at each token position, we can test the hypothesis that spelling and grammatical errors are detectable by aggregating these distributions across the entire text. For the experiments described below, we utilize the multilingual Qwen2.5-3B model. 

The following values are calculated at each step in token sequence given the discrete probability distribution \[ P = \{p_1, p_2, ..., p_V\} \] at the fixed time step, where $V$ is a vocabulary size of the LM tokenizer.

\textbf{Token probability} $p_t$ at the time step where $t$ is the token index.

\textbf{Entropy}
\begin{equation}
  \label{eq:entropy}
  H = -\sum_{i=1}^{V} p_i * log\;p_i
\end{equation}

\textbf{Entropy delta}
\begin{equation}
  \label{eq:entropy_delta}
  \delta\:H = -log\;p_t - H
\end{equation}

\textbf{Number of possible states} 
\begin{equation}
  \label{eq:num_possible_states}
  \eta = [e^H]
\end{equation}
where $[x]$ denotes integer part of $x$.

\textbf{Cumulative probability of possible states}
\begin{equation}
  \label{eq:cum_proba}
  \pi = \sum_{j=1}^{\eta} desc(P)_j
\end{equation}
where desc(P) stands for sorting $p_1,... p_V$ in descending order.

\textbf{Token rank} $r_t$ is an index of the token $t$ in the array $desc(p_1,... p_V)$

\textbf{Oddballness} $\xi$ is introduced by \citet{gralinski2024oddballness}:
\begin{equation}
  \label{eq:oddballness}
  \xi = \sum_{j=1}^{V} (p_j - p_t)^+
\end{equation}
where \[ x^+ = max(0, x) \]

An example of what an linguistic anomaly in text looks like at the token level can be seen in the Table \ref{tab:token_level_twinkle}. The token chain is the second line of "Twinkle, twinkle little star"~\footnote{~\url{https://en.wikipedia.org/wiki/Twinkle,_Twinkle,_Little_Star}}, in which the word "I" is changed to "It". The right half of the table reflects the change in metrics as a result of such a distortion. All probabilities of tokens following "It" are diminished, entropy values are increased.

\begin{table*}[h!]
\centering
\begin{tabular}{llcccclcccc}
\toprule

 & \multicolumn{5}{c}{Original line} & \multicolumn{5}{c}{Distorted line} \\
\cmidrule(lr){2-6} \cmidrule(lr){7-11}
$i$ & Token & $p_t$ & $r_t$ & $H$ & $\eta$ & Token & $p_t$ & $r_t$ & $H$ & $\eta$ \\
\hline
 10 & How      & 5.56e-01 & 1           &  3.398                & 29 & How      & 5.56e-01 & 1           &  3.398                & 29 \\
 11 & \textbf{ I}       & \textbf{9.74e-01} & \textbf{1}           &  \textbf{0.210}                & \textbf{1}  &  \textbf{ It}      & \textbf{6.56e-07} & \textbf{926}         &  \textbf{0.210}                & \textbf{1} \\
 12 &  wonder  & 9.81e-01 & 1           &  0.172                & 1  &  wonder  & 3.40e-06 & 5938        &  5.430                & 228 \\
 13 &  what    & 9.85e-01 & 1           &  0.130                & 1  &  what    & 1.77e-02 & 9           &  4.344                & 76 \\
 14 &  you     & 9.93e-01 & 1           &  0.075                & 1  &  you     & 4.31e-01 & 1           &  3.240                & 25 \\
 15 &  are     & 9.87e-01 & 1           &  0.100                & 1  &  are     & 8.51e-01 & 1           &  0.902                & 2 \\
 16 & !        & 5.65e-01 & 1           &  1.957                & 7  & !        & 3.40e-01 & 1           &  2.955                & 19  \\
 
\bottomrule
\end{tabular}
\caption{Token-level metrics (token probability $p_t$, token rank $r_t$, mean entropy $H$ and number of possible states $\eta$) for second line of L. Caroll's 'Twinkle, twinkle ...' poem.}
\label{tab:token_level_twinkle}
\end{table*}

The token-level values listed above are aggregated using min, max and median functions for get the text-level features.

\begin{figure}[h!]
\includegraphics[width=0.5\textwidth]{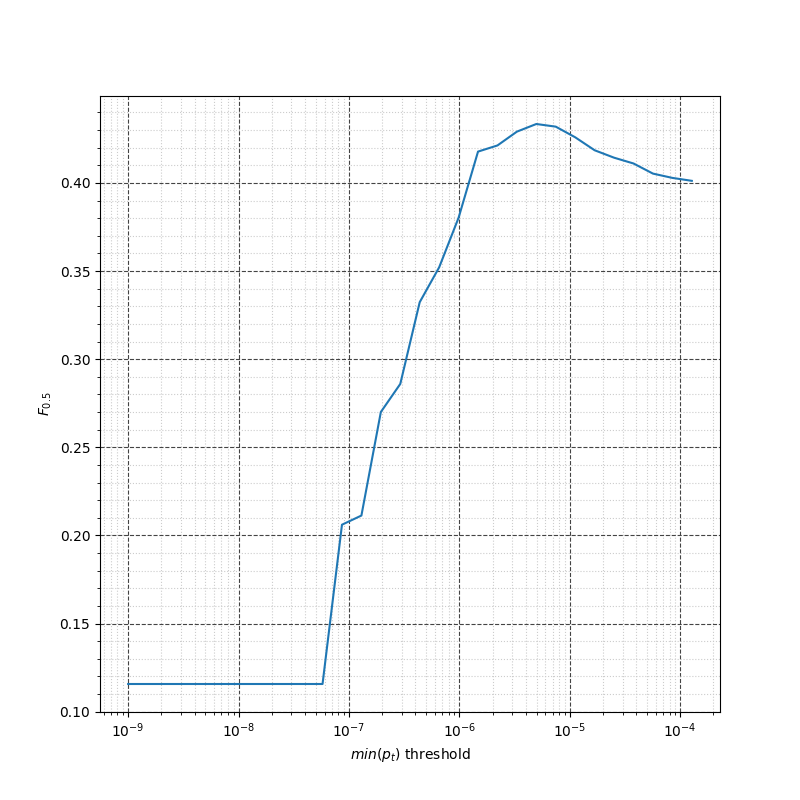}
\caption{Grid search of $p_t$ threshold and corresponding $F_{0.5}$ metric on the \texttt{RUPOR poetry}.}
\label{fig:gridsearch_threshold_token_p}
\end{figure}

\begin{figure}[h!]
\includegraphics[width=0.5\textwidth]{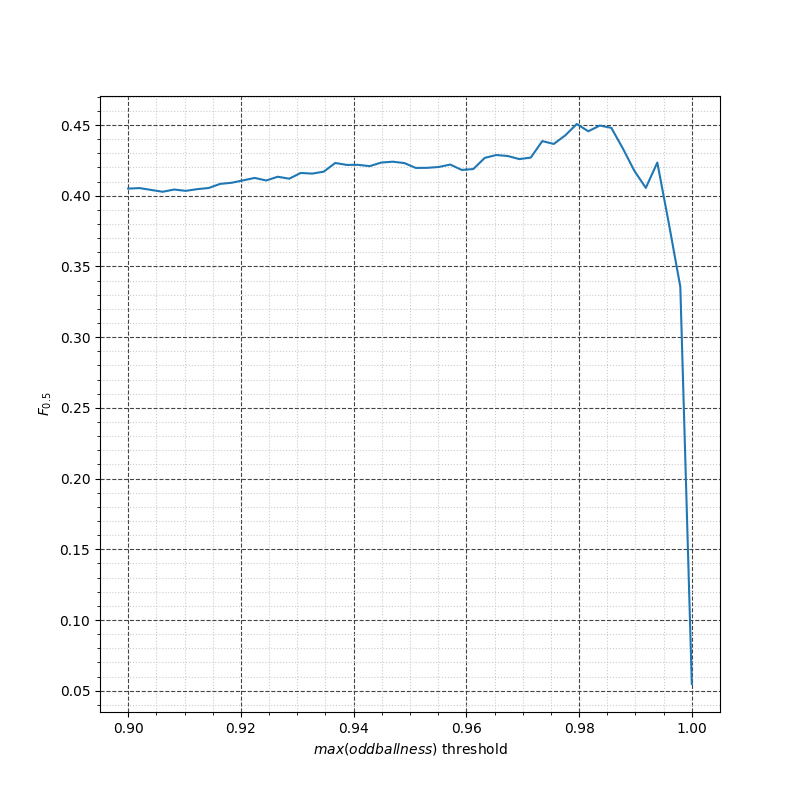}
\caption{Grid search of $\xi$ threshold and corresponding $F_{0.5}$ metric on the \texttt{RUPOR poetry}.}
\label{fig:gridsearch_threshold_oddballness}
\end{figure}

The simplest approach to detecting anomalies based on the collected features is to set a threshold below which \(min(p_t)\) will indicate the presence of at least one anomalous token. For other features, for example $\xi$, such a signal will be exceeding the specified threshold. The threshold can be selected through grid search and calculating the classification metric. Unfortunately, our experiments showed extremely low efficiency of this approach even in comparison with outlier detection approach described in Section~\ref{sec:outliers_detection}. The best $F_{0.5}$ value for $max(\xi)$ is 0.45. This is slightly higher than 0.43 for $min(p_t)$, but clearly not enough to detect poems with linguistic anomalies well. The Figures~\ref{fig:gridsearch_threshold_token_p} and \ref{fig:gridsearch_threshold_oddballness} show the corresponding dependencies of $F_{0.5}$ on the threshold value.

\begin{figure}[h!]
\includegraphics[width=0.5\textwidth]{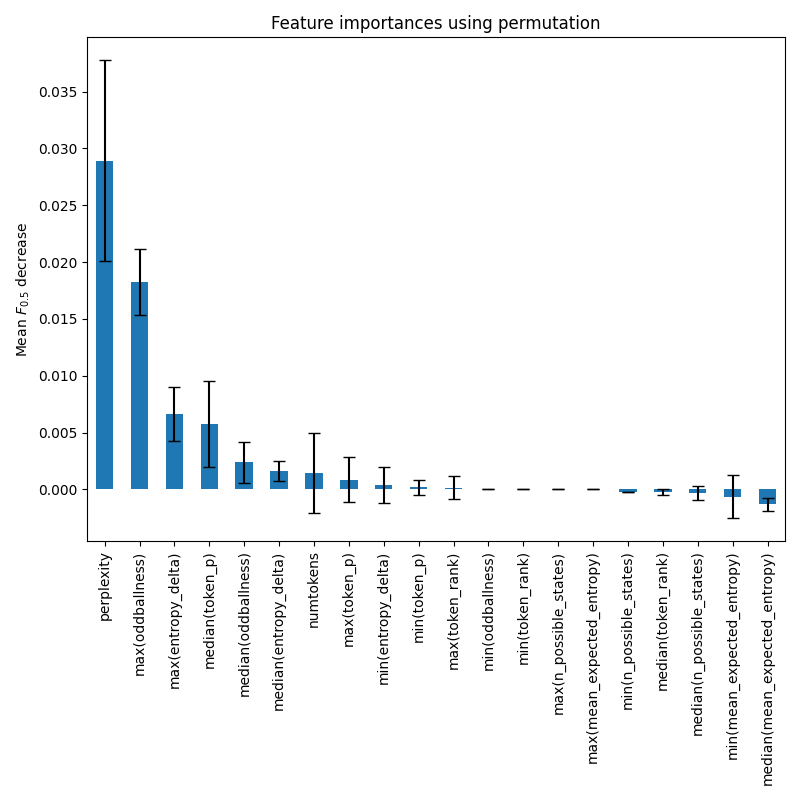}
\caption{Permutation feature importance at the text level for a binary classifier in the linguistic anomaly detection task.}
\label{fig:token-level-feature-importance}
\end{figure}

Instead of selecting a threshold value for individual features, you can use them all as features for a classifier that should detect texts with anomalies. This will allow you to estimate feature importance. Unfortunately, even in this setting, the achieved detection quality is very low. The importance of individual features, estimated using the permutation\_importance~\footnote{\url{https://scikit-learn.org/stable/modules/generated/sklearn.inspection.permutation_importance.html}} and the GradientBoostingClassifier~\footnote{\url{https://scikit-learn.org/stable/modules/generated/sklearn.ensemble.GradientBoostingClassifier.html}} from scikit-learn, is shown in Figure~\ref{fig:token-level-feature-importance}. Most important features include perplexity and oddballness.

\subsection{Results for Zero-Shot Detection of Linguistic Anomalies}
\label{app:zero-shot-results}

Mean $F_{0.5}$ and confidence intervals calculated at a 95\% confidence level using bootstrap resampling with 1,000 iterations for the \texttt{RUPOR} dataset are presented in Table~\ref{tab:zeroshot_llm_all_results}.

\begin{table*}[h!]
\centering
\begin{tabular}{lcccccccc}
\toprule
 & \multicolumn{2}{c}{\texttt{RUPOR} poetry} & \multicolumn{2}{c}{\texttt{RUPOR} prose} & \multicolumn{2}{c}{MultidomainGold} & \multicolumn{2}{c}{RUSpellRU} \\
\cmidrule(lr){2-3} \cmidrule(lr){4-5} \cmidrule(lr){6-7} \cmidrule(lr){8-9}         
\textbf{Model/Service} & $\overline{F_{0.5}}$ & 95\% CI & $\overline{F_{0.5}}$ & 95\% CI & $\overline{F_{0.5}}$ & 95\% CI & $\overline{F_{0.5}}$ & 95\% CI \\
\hline
YandexSpeller & \textbf{0.77} & 0.73, 0.80 & \textbf{0.78} & 0.75, 0.81 & \textbf{0.85} & 0.82, 0.87 & \textbf{0.83} & 0.81, 0.86 \\
GigaChat-20B-instruct & 0.57 & 0.52, 0.61 & 0.70 & 0.66, 0.74 & 0.66 & 0.62, 0.70 & 0.76 & 0.72, 0.79 \\
T-lite-it-1.0 & 0.58 & 0.55, 0.61 & 0.60 & 0.57, 0.64 & 0.58 & 0.54, 0.61 & 0.59 & 0.56, 0.63 \\
Qwen2.5-3B-Instruct & 0.39 & 0.33, 0.45 & 0.60 & 0.57, 0.64 & 0.58 & 0.54, 0.61 & 0.69 & 0.65, 0.73 \\
Mistral-7B-Instruct & 0.55 & 0.52, 0.58 & 0.55 & 0.52, 0.59 & 0.56 & 0.53, 0.60 & 0.55 & 0.52, 0.58 \\
\bottomrule
\end{tabular}
\caption{Mean $F_{0.5}$ scores, confidence interval (CI) lower and upper bounds for spellcheckers and zero-shot instruction-tuned LLMs on the \texttt{RUPOR} and \texttt{SAGE} corrupted text detection.}
\label{tab:zeroshot_llm_all_results}
\end{table*}

\subsection{Binary Classifiers Evaluation Setup and Results}
\label{app:binary-classifiers}

Foundation language models were fine-tuned for one epoch using a linear learning rate schedule on the training split of the combined dataset described in Section~\ref{sec:data}. The task involved binary classification of input text as either \textit{corrupted} or \textit{correct}. For encoder-only and decoder-only models, the classifier was implemented using the \texttt{AutoModelForSequenceClassification} class from the \texttt{transformers} library~\footnote{\url{https://huggingface.co/docs/transformers/model_doc/auto\#automodelforsequenceclassification}}. For the encoder-decoder T5 architecture, a sequence-to-sequence (seq2seq) approach was employed, where the input text was fed into the model, and the output text was either "True" or "False" to indicate the classification result.

The experiments were conducted using the \texttt{transformers} library (v4.48.0)~\footnote{\url{https://github.com/huggingface/transformers}}, CUDA v12.6, and 7 NVIDIA A100 GPUs. Fine-tuning hyperparameters, including the learning rate, were tuned for each model to minimize gradient norm spikes during training. The fine-tuning process required approximately 1 hour for ruRoberta and rugpt3medium and up to 8 hours for Qwen2.3-3B. Detailed hyperparameters are provided in Table~\ref{tab:binary-classifiers-lr}. LoRa settings are r=64, dropout=0.1, alpha=128.

The Python code for supervised fine-tuning, along with complete evaluation logs, is available in the repository~\ref{app:repository}.

\begin{table*}[h!]
\centering
\begin{tabular}{lcccc}
\toprule
\textbf{Foundation model} & num. of trainable params & learning rate & optimizer & precision \\
\hline

ruRoberta-large & 355M & 1e-5 & adamw\_torch & fp16 \\
rugpt3medium & 356M & 1e-5 & adamw\_torch & bf16 \\
FRED-T5-1.7B & 1663M & 5e-6 & adafactor & bf16 \\ 
Qwen2.5-3B & 3086M & 1e-6 & adamw\_torch & bf16 \\
Qwen2.5-3B LoRa & 14M & 1e-5 & adamw\_torch & bf16 \\

\bottomrule
\end{tabular}
\caption{Hyper-parameters of LM-based binary classifiers fine-tune.}
\label{tab:binary-classifiers-lr}
\end{table*}

After fine-tuning, the classifiers were evaluated on the test split of the combined dataset (see Section~\ref{sec:data}). To ensure fair comparison of metrics across domains, we balanced the classes in each domain through undersampling. The resulting sample sizes for each test domain after balancing are reported in Table~\ref{tab:binary-classifiers-test-population}.

\begin{table}[h!]
\centering
\begin{tabular}{lc}
\toprule
\textbf{Domain} & \text{Num. of samples} \\
\hline
RUPOR prose & 21999 \\
RUPOR poetry & 13595 \\
SAGE MultidomainGold & 2440 \\
SAGE RUSpellRU & 1302 \\

\bottomrule
\end{tabular}
\caption{Number of samples in test domains after class balance adjustment.}
\label{tab:binary-classifiers-test-population}
\end{table}

To ensure robustness, each model was fine-tuned and evaluated three times, allowing us to calculate 95\% confidence intervals for the $F_{0.5}$ scores. The evaluation results are reported in Table~\ref{tab:binary-classifiers-full-results}.

\begin{table*}[h!]
\centering
\begin{tabular}{lcccccccc}
\toprule
 & \multicolumn{2}{c}{\texttt{RUPOR} poetry} & \multicolumn{2}{c}{\texttt{RUPOR} prose} & \multicolumn{2}{c}{MultidomainGold} & \multicolumn{2}{c}{RUSpellRU} \\
\cmidrule(lr){2-3} \cmidrule(lr){4-5} \cmidrule(lr){6-7} \cmidrule(lr){8-9}         
\textbf{Model/Service} & $\overline{F_{0.5}}$ & 95\% CI & $\overline{F_{0.5}}$ & 95\% CI & $\overline{F_{0.5}}$ & 95\% CI & $\overline{F_{0.5}}$ & 95\% CI \\
\hline

FRED-T5-1.7B & \textbf{0.863} & 0.857, 0.87 & \textbf{0.825} & 0.819, 0.831 & \textbf{0.892} & 0.889, 0.896 & \textbf{0.946} & 0.944, 0.948  \\

ruRoberta-large & 0.802 & 0.787, 0.818 & \textbf{0.825} & 0.821, 0.829 & 0.882 & 0.862, 0.902 & 0.936 & 0.921, 0.951\\

Qwen2.5-3B & 0.810 & 0.804, 0.816 & 0.767 & 0.754, 0.78 & 0.843 & 0.817, 0.869 & 0.914 & 0.890, 0.938\\

Qwen2.5-3B LoRa & 0.749 & 0.710, 0.789 & 0.778 & 0.766, 0.791 & 0.835 & 0.819, 0.850 & 0.910 & 0.897, 0.924 \\

rugpt3medium & 0.615 & 0.535, 0.694 & 0.767 & 0.744, 0.790 & 0.837 & 0.797, 0.877 & 0.908 & 0.875, 0.942 \\

\bottomrule
\end{tabular}
\caption{$F_{0.5}$ scores for supervised binary classifiers on corrupted text detection. Confidence intervals (CI) are calculated at a 95\% confidence level with a t-value of 4.3.}
\label{tab:binary-classifiers-full-results}
\end{table*}

\end{document}